%% file: main.tex
\documentclass[10pt,twocolumn,letterpaper]{article}

\usepackage[pagenumbers]{cvpr} 


\definecolor{cvprblue}{rgb}{0.21,0.49,0.74}
\usepackage[pagebackref,breaklinks,colorlinks,allcolors=cvprblue]{hyperref}
\usepackage{graphicx}
\usepackage{caption} 
\usepackage{subcaption}

\usepackage[utf8]{inputenc} 
\usepackage[T1]{fontenc}    
\usepackage{hyperref}       
\usepackage{booktabs}       
\usepackage{amsfonts}       
\usepackage{nicefrac}       
\usepackage{microtype}      
\usepackage{xcolor}         
\usepackage{graphicx}
\usepackage{amsmath}
\usepackage{enumitem}
\usepackage{tabularx}
\usepackage{bm}
\usepackage{makecell}
\usepackage{booktabs}
\usepackage{float}

\def\paperID{20316} 
\def\confName{CVPR}
\def\confYear{2026}

\title{Few Shots Text to Image Retrieval: New Benchmarking Dataset and Optimization Methods}

\author{
    Ofer Idan\thanks{Emails: \{ofer.idan, vladi.vexler, gil.lederman, dima.sivov, aviad.cohen.zada1, shir.niego\}@huawei.com} \and 
    Vladi Vexler \and Gil Lederman \and 
    Dima Sivov \and Aviad Cohen Zada \and Shir Niego Komforti \\
    Huawei Tel-Aviv Research Center
}

\begin{document}
\maketitle

\begin{abstract}
Pre-trained vision-language models (VLMs) excel in multimodal tasks, commonly encoding images as embedding vectors for storage in databases and retrieval via approximate nearest neighbor search (ANNS). However, these models struggle with compositional queries and out-of-distribution (OOD) image-text pairs. Inspired by human cognition's ability to learn from minimal examples, we address this performance gap through few-shot learning approaches specifically designed for image retrieval.
We introduce the Few-Shot Text-to-Image Retrieval (FSIR) task and its accompanying benchmark dataset, FSIR-BD—the first to explicitly target image retrieval by text accompanied by reference examples, focusing on the challenging compositional and OOD queries. The compositional part is divided to urban scenes and nature species, both in specific situations or with distinctive features. FSIR-BD contains 38,353 images and 303 queries, with 82\% comprising the test corpus (averaging per query 37 positives, ground truth matches, and significant number of hard negatives) and 18\% forming the few-shot reference corpus (FSR) of exemplar positive and hard negative images.
Additionally, we propose two novel retrieval optimization methods leveraging single shot or few shot reference examples in the FSR to improve performance. Both methods are compatible with any pre-trained image encoder, making them applicable to existing large-scale environments. Our experiments demonstrate that: (1) FSIR-BD provides a challenging benchmark for image retrieval; and (2) our optimization methods outperform existing baselines as measured by mean Average Precision (mAP). Further research into FSIR optimization methods will help narrow the gap between machine and human-level understanding, particularly for compositional reasoning from limited examples.
\end{abstract}


\section{Introduction}\label{sec:intro}

The rapid increase of smart city infrastructure, CCTV networks, autonomous driving systems, and data centers has created an unprecedented demand for continuous video and image processing at scale. These applications require systems that can handle constant data streams while maintaining low latency, minimizing costs, and maximizing throughput. The challenge is compounded by the diverse nature of visual queries: some involve familiar domains but require understanding of compositionally complex concepts—such as identifying "uncovered manholes" or detecting "illegal occupation of public space by private stores" in urban management contexts. Others venture into specialized territories like medical imaging or satellite analysis, where visual concepts may be entirely unfamiliar. Across all scenarios, the stakes are high: accuracy cannot be compromised, yet collecting comprehensive training datasets for fine-tuning remains prohibitively expensive and time-consuming.

At the heart of these challenges lies a fundamental question: how can we efficiently search vast image databases with natural language queries? Foundational Visual Language Models (VLMs) provide an elegant answer by learning joint representations that map both text and images into a shared semantic space. This approach enables intuitive similarity-based retrieval, where systems store precomputed image embeddings in vector databases like FAISS and identify the most semantically similar matches to text query embeddings through simple cosine distance calculations.

However, VLMs are far from perfect. While they excel at recognizing frequent concepts well-represented in their training data, they falter when confronted with rare examples, compositional reasoning tasks, logical operations like negation, or nuanced relationships between concepts. When users encounter these limitations in practice, they instinctively adapt—marking a few relevant results from initial searches and using these examples to refine their queries. This natural human response reveals an important opportunity: the Few-Shot Text-Image Retrieval (FSIR) problem, where a handful of visual examples can dramatically improve search performance.

Our investigation of smart city deployments revealed a compelling architectural insight. These systems follow a consistent two-phase pattern: during the understanding phase, cameras and field personnel continuously stream visual data to centralized databases, where computational efficiency is paramount. Here, lightweight dual-encoder architectures handle the heavy lifting of image vectorization and storage. But during the search phase, the user base shrinks dramatically—primarily consisting of operators and analysts—creating a computational bottleneck that can accommodate more sophisticated processing. This asymmetry presents a strategic opportunity: we can deploy powerful Multimodal Large Language Models (MLLMs) exclusively during search operations, interfacing them cleverly with the simpler encoders used for data ingestion.

Building on these insights, we make several key contributions to advance the field. We begin by formalizing Few-Shot Text-Image Retrieval as a distinct research problem and systematically analyzing the limitations of existing approaches. We then introduce two innovative solutions: first, a prompt learning method that harnesses few-shot visual examples to automatically generate query-specific prompts, and second, a general-purpose MLLM approach that learns to align with external image encoder embedding spaces, enabling dynamic generation of search embeddings from text queries and reference images without requiring prior exposure to specific concepts. To facilitate rigorous evaluation, we contribute FSIR-BD, a carefully curated benchmark dataset for Composed queries of Text plus References (CTR). Our experimental results demonstrate that both proposed methods achieve state-of-the-art performance, substantially outperforming existing baselines across diverse scenarios.

The practical implications of FSIR extend well beyond theoretical advances. The paradigm enables rapid, query-specific performance improvements without the traditional overhead of foundational model retraining or fine-tuning. The process is elegantly simple: users collect just one or a few examples per query—often by marking correct results from initial top-k retrievals or identifying near-miss hard negatives that highlight subtle distinctions. These reference images then guide our FSIR-PL or FSIR-CTR methods (and hopefully future innovations from the broader research community) to generate more precise search representations.

This approach has already proven its worth in production environments. FSIR-PL currently serves as a core component of a cloud service for video and image semantic management, successfully deployed across multiple large-scale enterprise scenarios. Our field experience confirmed a crucial insight: while fine-tuned encoders provided modest baseline improvements, FSIR methods enabled rapid, targeted enhancements that could be applied immediately to address specific query challenges, without the need to re-encode existing datasets. This real-world validation motivated our commitment to developing the FSIR benchmark and sharing it with the research community, fostering continued innovation in this promising domain.

To summarize, our contributions are as follows:
\begin{itemize}
    \item We motivate and formally define the Few-Shot Text-Image Retrieval task.
    \item We publish FSIR-BD, a CTR benchmark with \emph{positive and hard negative samples} and \emph{multiple} correct images per query, spanning urban life, natural world, and OOD queries.
    \item We introduce open-query plus few-shots prompt-learning method, the FSIR-PL, and demonstrate its high efficiency compared to the baseline.
    \item We present FSIR-CTR, enabling scaling up a training dataset (or small-dataset training) and learning new concepts from text and reference images as searchable embeddings aligned to any image encoder's latent space. We demonstrate its efficiency against the baseline.   
\end{itemize}

\section{Related Work}

\subsection{Vision Language Foundation Models}
Vision language models~\citep{radford2021learning, ALBEF} pre-train image-text encoders on large-scale data containing hundreds of millions of image-caption pair. Following CLIP, many vision-language “foundation” models incorporate more data into training, introduce new architectural designs, or utilize new objective~\citep{li2022blip}.
VLM pre-training learns correlate between image and text using large scale image-text pairs, which enables zero shot retrieval on visual recognition tasks. Given image-text pairs, Image encoder and text encoder extract features and then learn visual-language correlation according to pre-training objectives.

\subsection{Weaknesses of VLMs}
As described above, VLMs are trained on image-caption tuples, and work by encoding text and image into the same latent space. They can be used as an open query Text-to-Image (T2I) and Image-to-Text (I2T) retrieval tool. However, they are known to have several weaknesses. One in particular is the challenge of \textit{Compositionality}, understanding the meaning of the whole from the meaning of the parts.

The issue of Compositionality is the name given to an aggregation of allegedly related problems in VLMs. They are much better at detecting objects than detecting either the relationship between or the attributes of objects~\citep{Thrush_2022, yuksekgonul2022visionlanguage}, have difficulties with fine-grained physical grounding of objects~\citep{Zhang2024CounterCurateEP}, and degrade in performance as known elements are composed in growing syntactic complexity~\citep{Ma_2023, hsieh2023sugarcrepe}. These difficulties of VLMs are exposed in the ``compositionality benchmarks'' by means of hard negative distractors, usually in text~\citep{Ma_2023}, but recently also in images~\citep{ray2023cola, mir2024bivlc}. Our compositionality benchmark, unlike previous ones, is a Text-to-Image dataset, where each query has dozens of positive answers.

\subsection{Composed Image Retrieval}
A method most similar to ours is Composed Image Retrieval (CIR) - where the model retrieves an image based on a reference image and instructions expressing user intent for modifying it. It is a challenging vision language (VL) task that takes a composed query of image and text, aiming to search relative images for both conditions \cite{saito2023pic2word, gu2024lincir, zhang2024magiclens, gu2023compodiff, gulanguage}. In contrast to the image-based retrieval systems \cite{radford2021learning, li2022blip}, CIR is advantageous as it allows retrieval of images with a higher precision thanks to the text query that incorporates user’s intent, such as a desired modification to the query image. Recently, MLLM were used for CIR task and achieved SOTA results \cite{jiang2024e5, jiang2024vlm2vec}. However, MLLM are costly and require large GPU memory and long processing time especially for encoding images which produce many tokens.

In~\cite{wu2019fashion} they introduced the FashionIQ dataset, which is a CIR dataset with 50k instances. The task is to retrieve an image of a dress that is similar to the reference image, but in a different color. The query is a text description of the desired modification (e.g., "Same dress, but in blue"). The CIRR dataset was introduced in~\cite{Liu_2021}, applying CIR to real world images (rather than narrow domains such as fashion), with 36k instances. In~\cite{Levy_2024}, authors introduce LaSCo, a large-scale CIR dataset, and the CASE baseline. They define the "modality redundancy" metric for CIR datasets, measuring how redundant the two modalities are (can the text of reference image be used for retrieval on their own?). The Zero-Shot CIR task was introduced in~\cite{Baldrati_2023}, along with the CIRCO dataset. They avoid the need for a large labeled dataset by means of a pre-training stage which learns to translate an unlabeled image into a single VLM (CLIP) "word" (a technique called ``textual inversion'') and, thereby concatenating it to the text and transforming the problem into standard T2I. CIRCO has around 1300 instances. ~\citep{saito2023pic2word} also uses the textual inversion technique to solve the ZS-CIR task. The problem of Few-shot CIR was formally defined in~\cite{Wu_Wang_Zhao_Zhang_Lu_Li_Henao_2023}, where the retriever is given a few labeled triplets, and the textual inversion method is used to enable prompt tuning. While CIR task uses text to apply changes on the reference image (such as [image]+"this dress but in red", as Fashion-IQ dataset), FSIR task emphasizes the shared part between the text and the reference image to increase retrieval accuracy.

\subsection{Robust fine-tuning}
One of the most appealing properties of foundational models is their robustness to OOD samples. However, fine-tuning foundation models to a new task is known to degrade their robustness to OOD samples~\citep{radford2021learning,Pham_2023,andreassen2021evolution}. This led to ongoing research on robust fine-tuning, aiming to achieve improved performance on the target distribution, while preserving the zero-shot capabilities of the foundation model~\citep{2022Mitchellarxiv:2203.05482,Goyal_2023,2023Jishnuarxiv:2308.13320}. 

The problem can otherwise be alleviated by utilizing ideas from parameter-efficient transfer learning~\citep{2019Neilarxiv:1902.00751}, specifically learning small adapters to larger models~\citep{2021Junxianarxiv:2110.04366,2021Pengarxiv:2110.04544,2021Yi-Linarxiv:2112.06825}, so that the original model remains intact. Our method similarly leaves the model intact, but rather uses the technique of Prompt Tuning.
\subsection{Few-Shot Prompt Learning}\label{fs-classification}
Few Shot Prompt Learning (FSPL) aims to learn language context for text encoder input, these tokens are trained with cross-modal training \cite{zhou2022coop, zhou2022cocoop, prograd2023}. They use a fixed CLIP text encoder and CLIP image encoder and learn a context using N shots examples of K classes.
This approach requires fewer samples, easing its use and cost-effectiveness. Being local, It also solves the catastrophic forgetting problem. It is used for N-way-K-shot classification, where all classes (“way”) are learned simultaneously, resulting in a single learned vector. This vector is combined, as a prefix, with class names to create final N prompt-vectors. For image classification, all vectors are used, and cosine similarity determines the class, with the highest score indicating the class and confidence level. In contrast to \cite{zhou2022coop, zhou2022cocoop, prograd2023} which target N-way classification task, and fail on classes not seen in the N-way training data, our approach targets retrieval with unlimited number of classes. In addition, it answer the following problems of FSPL: 1) How to perform a MM Prompt-Learning with K-shots and One-Way (1 query/prompt) fitting a retrieval scenario (vs. previous scenarios dealing with classification or generative models and scenarios). 2) Introduce Learning Loss methods required for MM Retrieval Prompt Learning. 3) Introduce a way to control the level of relevance between multimodal items, such as (but not limited to): a) hard positives (HP), b) hard-negatives (HN), and c) unrelated easy negatives (EN)

\section{FSIR Benchmarking Dataset (FSIR-BD)}
Contemporary image retrieval systems face challenges that existing benchmarks inadequately address. Real-world scenarios involve users submitting concise queries targeting specific image aspects, with compositional elements like object attributes, relationships, and negation constraints. Expected results comprise multiple relevant items ranked by relevance, differing substantially from traditional image-caption paired datasets. We identify four critical limitations in existing benchmarking approaches:

\begin{itemize}

\item \textbf{Compositionality and conciseness of queries.} Widely-used datasets such as COCO, Flickr30K, and ImageNet-1k \cite{lin2014microsoft, plummer2015flickr30k, deng2009imagenet} employ one-to-one image-caption mapping with captions emphasizing semantic main aspects. This poorly reflects real-world usage, where users employ brief, targeted queries focusing on specific components—inherently compositional and including attributes, spatial relationships, long-tailed aspects, and negation constraints that jointly filter or promote relevant items.

\item \textbf{Result cardinality.} Traditional datasets emphasize singular correct matches, contrasting with practical applications where multiple results may satisfy queries with varying relevance. CIRCO \cite{Baldrati_2023} addresses this by providing multiple ground truths per query.

\item \textbf{Lack of Positive Examples with Hard Negatives.} CIR datasets such as CIRCO \cite{Baldrati_2023} and CIRR \cite{CIRR_Liu_2021_ICCV} provide a reference image paired with modification text, where the reference serves as a hard negative, but lack additional positive examples. 

\item \textbf{Performance metrics.} Prevalent Recall@K metrics, suitable for one-to-one scenarios, fail to assess ranking quality. Mean Average Precision (mAP@K) better reflects real-world requirements by incorporating both recall and ranking quality, as emphasized in CIRCO.
\end{itemize}

To our knowledge, FSIR-BD is the first dataset addressing all above crucial points.

\begin{figure}
    \centering    
    \begin{subfigure}[b]{0.3\textwidth}
        \includegraphics[scale=0.7, width=\textwidth]{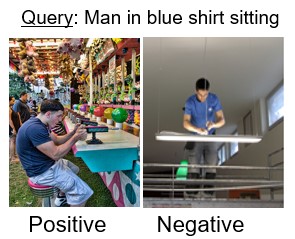}
        \caption{Compositional-VG}
    \end{subfigure}
    \hfill
    \begin{subfigure}[b]{0.32\textwidth}
        \includegraphics[scale=0.7,width=\textwidth]{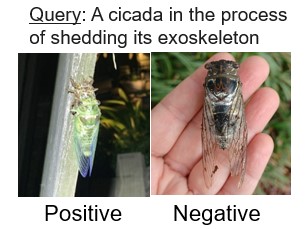}
        \caption{Compositional-INQUIRE}
    \end{subfigure}
    \hfill
    \begin{subfigure}[b]{0.35\textwidth}
        \includegraphics[scale=0.7,width=\textwidth]{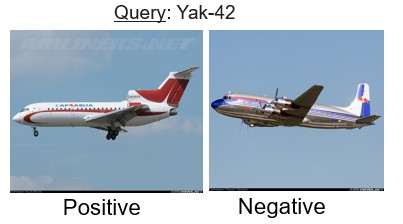}
        \caption{OOD}
    \end{subfigure}
    \caption{Example of FSIR-BD. Per query text we have multiple positive and hard-negative images.}
    \label{fig:fsir_bd}
\end{figure}

\subsection{FSIR-BD Sub-datasets}
Our dataset comprises three components: \textbf{Urban life} (FSIR-BD-Compositional-VG), \textbf{Natural world} (FSIR-BD-Compositional-INQUIRE), and \textbf{Out-of-Distribution} (FSIR-BD-OOD). Together, they span diverse domains relevant to practical applications. The following sections describe their construction.

\textbf{General aspects}: We aimed for $\sim$50 hard positives (HP) per query, up to 3$\times$ as many hard negatives (HN), plus easy negatives, forming the ground-truth query result set (GTQR). GTQR was split into a test set and a few-shot reference corpus (FSR), each query having 16 positives and 16 HNs in the FSR. This design balances statistical robustness, retrieval difficulty, and sufficient few-shot references.  

\subsubsection{FSIR-BD-Compositional-VG - Urban-life}
This dataset is based primarily on Visual Genome (VG) \cite{krishnavisualgenome} and supplemented with 1,200 images from Flickr32K \cite{plummer2015flickr30k}, it leverages VG's comprehensive collection of 108,077 images to provide the diversity and repeatability required for a compositional, multiple images per query, retrieval dataset. The construction, curation and cleanup of this sub-dataset demanded the most substantial effort, which is described in the relevant section in the Appendix.

The result is a corpus with 9,073 total images and 50 queries. Examples of our benchmark dataset in Figure \ref{fig:fsir_bd} (more examples available in our supplementary materials).

\subsubsection{FSIR-BD-Compositional-INQUIRE}
FSIR-BD-Compositional-INQUIRE construction was fully automated, based on the INQUIRE benchmark \cite{vendrow2024inquire} and 1,200 Flickr32K images. 
INQUIRE provides 250 expert-level retrieval queries over iNaturalist 2024 (5M images, 10K species), covering species, context, behavior, and appearance (e.g., \textit{"bird perched on a hippo"}), with 33,000 labeled matches. 

For FSIR-BD, we selected 97 test queries with at least 50 positives, randomly taking 50 positives per query and 150 HN images (same species), yielding a GTQR of 20,230 images.

\subsubsection{FSIR-BD-OOD }
This dataset construction was fully computerized, based on three established OOD databases: FGVC-Aircraft \cite{maji13fine-grained} with labeled aircraft model images, DTD \cite{cimpoi14describing} with labeled texture images, and EuroSAT \cite{helber2019eurosat}, \cite{helber2018introducing} with Sentinel-2 satellite images classified by land use and cover. 
For our FSIR-BD-OOD dataset, we randomly selected 50 positive samples from each class across these datasets. The resulting GTQR contains 9,050 images spanning 157 queries.

\subsection{Analysis and Quality Assessment}
We estimate up to 1\% false positives and 2\% false negatives in Compositional-VG and Compositional-INQUIRE, corresponding to F1-scores of 96.5 and 95.4—sufficiently robust for evaluation. In the OOD case, with single-class entities, we assume no labeling errors.  Following prior work \cite{2019Suhr-Arxiv:1811.00491, CIRR_Liu_2021_ICCV, CIRCO_baldrati2023zeroshot}, in Table~\ref{tab:FSIR-BD-Quant} we provide quantitative statistics for FSIR-BD and compare it with  CIRCO \cite{Baldrati_2023} and CIRR \cite{CIRR_Liu_2021_ICCV} datasets. Additional semantic coverage of the FSIR-BD is assessed in Appendix Table~1.  

Unlike CIR datasets, which use verbose edits and a single hard-negative, FSIR-BD provides short, self-contained queries paired with positive and 16 hard negative references. It spans diverse domains, emphasizing OOD, compositional, and “expert” queries. Queries include multiple positives\footnote{Multiple ground truths enable stable, fine-grained evaluation.} and 3$\times$ as many HN —properties absent from prior datasets and closer to real-world conditions.

\begin{table*}[h]
  \small
  \scalebox{0.85}{
    \centering
    \begin{tabular}{lccc}
        \toprule        
                            & FSIR     & CIRR   & CIRCO   \\                
        \midrule        
        Domains             & 3 (Urban-life, Natural world, OOD)   &  1 (Urban-life)   & 1 (Urban-life) \\
        Data sources        & Visual Genome, Inquire iNat24, FGVC-Aircraft, DTD, EuroSAT   &  NLVR2   & COCO 2017 Unlabeled  \\
        Nb. of Images total             & 38,353 & 21,552 & 123,000 \\
        Nb. of Images in test    & 31,449 & 3,315  & 123,000 \\
        Nb of Queries total             & 303    & 36,554 & 1,020 \\
        Nb of Queries in test    & 303    & 4,131  & 800 \\
        Queries type                    & Complete query & Modification text & Modification text \\
        Queries average sentence length & 5.96  & 11.3 & 10.4 \\
        References Positives per query  & 16     & 0      & 0 \\
        References Hard Negatives per query & 16 & 1      & 1 \\
        Ground Truths per query in test        & 37     & 1      & 4.53 \\
        Hard Negatives per query in test  & 110 & - & - \\ 
        Evaluation method               & mAP@K  & Recall@K & mAP@K \\
        
        \bottomrule
    \end{tabular}}
    \caption{Analysis of quantitative and benchmarking aspects of FSIR-BD, CIRR and CIRCO datasets.}
    \label{tab:FSIR-BD-Quant}
\end{table*}

\textbf{In Summary:} FSIR-BD is the first high-quality dataset for image retrieval by Composed Text+Reference (CTR) queries, emphasizing challenging compositional and OOD cases. It contains 38,353 images and 303 queries: 82\% in the test corpus (avg.\ 37 ground truths per query with up to 3$\times$ hard negatives) and 18\% in the FSR corpus.

\section{FSIR Retrieval Optimization Methods}

\subsection{K-Shot One-Way Prompt Learning (FSIR-PL)}
As discussed in Section~\ref{fs-classification}, existing VLM-based few-shot prompt learning methods address N-way k-shot classification by learning either shared prefixes across N class prompts or N class-specific prefixes. A naive adaptation of these methods (specifically,~\cite{zhou2022coop}) to FSIR would set N=2, treating retrieval as binary classification with relevant images as positives and non-relevant images as negatives.

This approach fails in the retrieval domain. It requires two distinct prompts per class: while the positive prompt can be "a photo of Q", the negative prompt must represent "a photo of not Q"—a fundamentally ill-defined concept spanning countless unrelated objects. VLMs struggle particularly with such negation.

\begin{figure}[ht!]
\centering
\includegraphics[width=85mm]{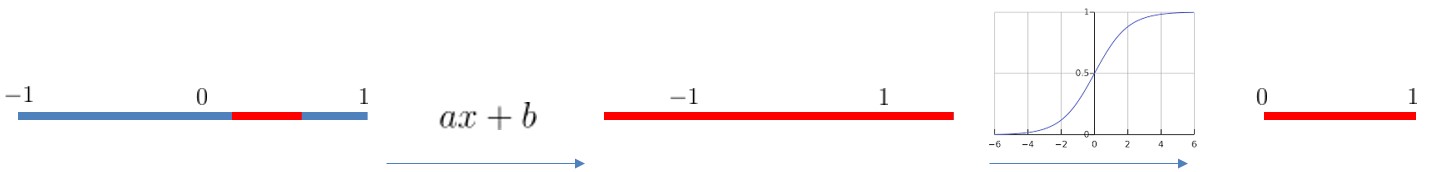}
\caption{BLIP scores fall within [0.1,0.5]; learnable scaling factors a,b map these to the full BCE range [0,1].}
\label{fig:pt_bce_loss}
\end{figure}

We propose learning a single prompt representing concept Q, trained with Binary Cross-Entropy (BCE) loss. Given query Q and an image, we compute cosine similarity between their learned text and image features, yielding score x. As shown in Figure~\ref{fig:pt_bce_loss}, we map this similarity to probability $p \in [0,1]$ using:

\begin{equation}
p = S(ax+b) = \frac{1}{1+e^{-(ax+b)}}
\end{equation}
where S is the sigmoid function, and learnable parameters a,b transform raw similarity scores from contrastive models like BLIP (typically [0.1,0.5]) into appropriate sigmoid inputs. Given probability p and binary label $y \in \{0,1\}$, the BCE loss is:

\begin{equation}
\mathcal{L}_{\text{BCE}} = -[y\log p + (1-y)\log(1-p)]
\label{eq:bce_loss}
\end{equation}

\begin{figure}[ht!]
\centering
\includegraphics[width=90mm]{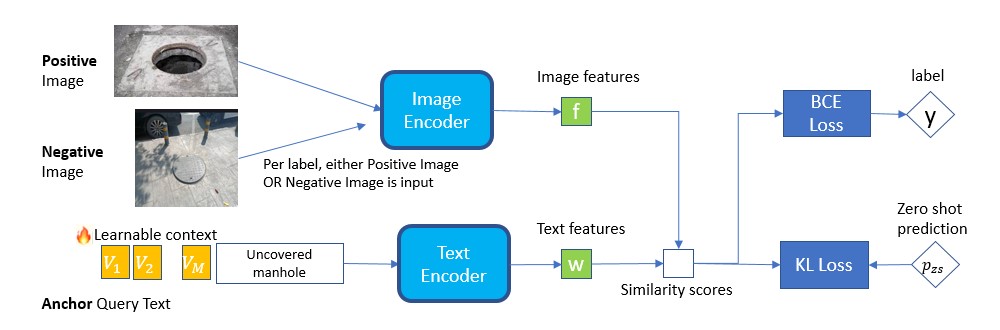}
\phantomsection
\caption*{Prompt learning training employs learnable vectors with positive-negative image pairs to optimize context vectors, establishing an effective decision boundary for 1-way retrieval tasks.}
\label{fig:pt_arch}
\end{figure}

Figure~\hyperref[fig:pt_arch]{1} presents the FSIR-PL architecture. The model uses 16 learnable context vectors concatenated with the user query, processed by the text encoder to generate text features. The image encoder processes positive or negative images (per class label) to produce image features. Cosine similarity scores x between text and image representations are transformed into probabilities via sigmoid for BCE loss calculation.

We incorporate Kullback-Leibler (KL) divergence loss~\citep{prograd2023}, quantifying the discrepancy between FSIR-PL score $p(\bm{t}_i|\bm{x})$ and zero-shot text encoder score $p_\text{zs}(\bm{w}_i|\bm{x})$. This loss is modulated by ProGrad factor $G_\text{prograd}$, which updates prompts whose gradients align with the zero-shot direction, preventing loss of VLM general knowledge:
\begin{equation}
    \mathcal{L}_\text{KL} = -\sum_i p_\text{zs}(\bm{w}_i|\bm{x}) \log \frac{p(\bm{t}_i|\bm{x})}{p_\text{zs}(\bm{w}_i|\bm{x})}
\end{equation}

The final objective combines BCE and KL divergence losses:

\begin{equation}
\mathcal{L} = \mathcal{L}_\text{BCE} + G_\text{prograd} \cdot \mathcal{L}_\text{KL}
\end{equation}

\subsection{Composed Text Plus Reference Shots Model (FSIR-CTR)} \label{sec:fsir_ctr}

\begin{figure}[ht!]
\centering
\includegraphics[width=90mm]{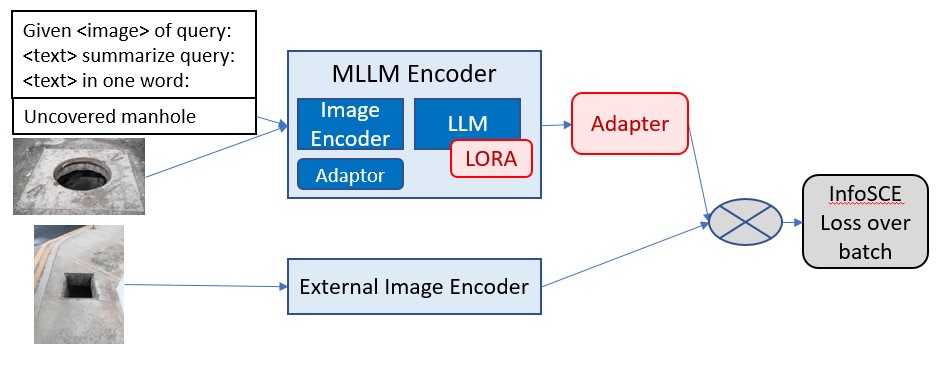}
\caption{FSIR-CTR training architecture. The reference image plus text query are input to the MLLM, which fuse them and align the result to the external image encoder's embedding space.}
\label{fig:vllm_cir_arch}
\end{figure}

The method of K-shots One-Way Prompt Learning requires per-query training, which proves computationally expensive and time-consuming for practical applications. To address this limitation, we propose a more efficient approach: the Composed Text Plus Reference Shots Model (FSIR-CTR). FSIR-CTR model accepts a combination of text and reference images (one or multiple) as input, generating a unified embedding vector. Figure~\ref{fig:vllm_cir_arch} illustrates the FSIR-CTR architecture.

The FSIR-CTR model comprises two principal components:

\noindent\textbf{External Image Encoder.} We employ a pre-trained visual backbone model to extract visual features from target images $y$, producing embeddings $v_{\phi}(y) \in \mathbb{R}^m$. We keep the weights of this external image encoder frozen during training, which facilitates rapid deployment across various image encoder architectures while preserving their original performance characteristics.

\noindent\textbf{Multimodal Large Language Model (MLLM) Encoder.} The MLLM encoder integrates information from both the target text query and the reference image to form a unified representation. This representation encodes a joint concept that combines the semantics of the query text with the visual content of the reference image (Figure~\ref{fig:vllm_cir_arch}). We employ LoRA~\cite{hu2022lora} to fine-tune the MLLM encoder to this task, alongside a linear projection layer, which aligns the MLLM's embedding space with that of the external pre-trained image encoder.

During training, we input the query instruction $q_{\text{inst}}$ comprising both query text and a reference image to the MLLM to obtain query embeddings $\mathbf{h}_{q_{\text{inst}}}$. These embeddings are generated by extracting the last-layer representation of the final token followed by passing it through a linear mapping layer. For $q_{\text{inst}}$, we use the prompt template: \textit{``Given <image> of query: <text>, summarize query: <text> in one word:''}, where \texttt{<image>} and \texttt{<text>} are replaced with the reference image and text query, respectively. Concurrently, we feed the target image to the external image encoder to generate the target embedding $\boldsymbol{v}$.

For model training, we utilize the standard InfoNCE loss to align the MLLM's query embedding with the external encoder's target image embedding through in-batch negatives (where t denotes the target):
\begin{equation} \label{equ:t2i_loss}
    \min \mathcal{L} = -\log \frac{\phi(\mathbf{h}_{q_\text{inst}}, \mathbf{v}_{t^+})}{\phi(\mathbf{h}_{q_\text{inst}}, \mathbf{v}_{t^+}) + \displaystyle\sum_{t^- \in \mathbb{N}}\phi(\mathbf{h}_{q_\text{inst}}, \mathbf{v}_{t^-})}
\end{equation}
where $\mathbb{N}$ denotes the set of all negative samples, and $\phi(\mathbf{h}_q, \mathbf{v}_t)$  quantifies the matching score between the query q representation from the MLLM and the target t
representation from the external image encoder. We employ a temperature-scaled cosine similarity function:
\begin{equation}
\phi(\mathbf{h}_q, \mathbf{v}_t) = \exp\left(\frac{1}{\tau}\cos(\mathbf{h}_q, \mathbf{v}_t)\right)
\end{equation}
where $\tau$ is a temperature hyperparameter.


During inference, we feed the query text along with its corresponding reference image into the MLLM to produce the query embedding. Although FSIR-CTR is trained with a single reference image, the MLLM architecture supports simultaneous processing of multiple reference images, thereby enabling improved adaptation to novel concepts.

\subsubsection{FSIR-CTR Training Dataset (FSIR-TD)}  \label{sec:FSIR-CTR-Training-dataset}
Training FSIR-CTR requires triplets of query text, reference image, and target image. We developed an automated pipeline to generate these triplets from image-caption pairs (COCO-Captions \cite{chen2015microsoft}).
The pipeline first encodes all images using BLIP \cite{li2022blip} and captions using ModernBert \cite{modernbert}, storing embeddings and metadata in a vector database. For each randomly sampled query text, we retrieve its matching query image (the target) and perform cosine similarity search to find the top-200 similar images. These candidates are re-ranked by comparing query text embeddings with their caption embeddings. Images with similarity > 0.65 are labeled as reference images (we attempted similar automatic labeling for hard negatives using lower thresholds, but found the results too noisy).

Figure \ref{fig:coco_db_1} shows examples from our training dataset (additional examples in supplementary material). Each reference image yields multiple target images, enabling dataset scaling and improved model generalization.


\begin{figure}[ht!]
\centering
\includegraphics[width=90mm]{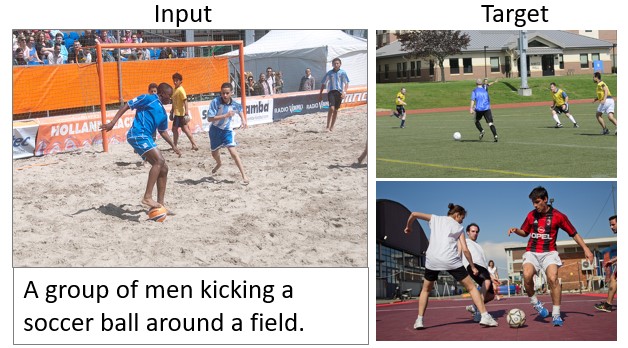}
\caption{Query text plus reference image vs. target images from FSIR-TD.}
\label{fig:coco_db_1}
\end{figure}

\subsubsection{FSIR-CTR Reference Image Selection Strategy}
During inference, we select reference images from a given pool to optimize FSIR-CTR performance. To assess various selection strategies, we utilized the FSIR-BD few-shot train-validation set as our reference image pool, which shares the same distribution as the test set.
Although FSIR-CTR was trained with single reference images, the MLLM architecture supports processing multiple reference images concurrently (at the cost of extended prompts, higher computational requirements, and increased latency). 
We identified the optimal reference image using the validation set and subsequently applied it to the test set for evaluation. Our selection algorithm operates in two stages: (1) we assess all individual reference images from the validation set (FSIR-BD FSR) and identify the one achieving the highest performance score, and (2) we combine multiple top-performing images based on their individual accuracies from step (1) and determine the best combination on the validation set. Additional implementation details are available in the supplementary materials. 

It should be noted that since the validation set (FSIR-BD FSR) includes only 16 hard positives and 16 hard negatives per query, selecting the highest-performing image on the validation set does not guarantee optimal test set performance. We observed better results using alternative selection strategies, suggesting that larger validation sets would improve optimal performance in large-scale deployments.


\section{Experimental Results}

\subsection{Experimental Setup}
We implemented two variants of FSIR with distinct architectural configurations. For FSIR-PL, we utilized CLIP~\cite{radford2021learning} and BLIP~\cite{li2022blip} as backbone vision-language models. We introduced a learnable prompt mechanism comprising 16 trainable context vectors, which were concatenated with the target query before being passed to the BLIP text encoder. The model was trained on FSIR-BD, where each query was paired with few-shot examples consisting of positive images, hard negatives, and unrelated easy negatives sampled from the Flickr32K dataset. Model was optimized using the Adam optimizer with a batch size of 32. Training was performed on a single NVIDIA V100 GPU.

In parallel with this approach, FSIR-CTR leveraged CLIP~\cite{radford2021learning} and BLIP~\cite{li2022blip} as external image encoders. The training procedure consisted of three sequential stages. First, we fine-tuned the generative Phi-3.5-V MLLM~\cite{phi35} to perform retrieval tasks, applying the E5-V approach~\cite{jiang2024e5} on text-only data. Second, we trained an alignment adapter to bridge the embedding spaces of the frozen fine-tuned Phi-3.5-V model (Stage 1) and a frozen external image encoder. Third, we conducted task-specific fine-tuning of the Phi-3.5-V model (Stage 1) for FSIR using the VLM2VEC training framework~\cite{jiang2024vlm2vec}, as described in Section~\ref{sec:fsir_ctr}. This progressive three-stage methodology enabled the model to specialize in each objective sequentially, yielding superior performance compared to single-stage training approaches.
FSIR-CTR model was optimized using the Adam optimizer with a batch size of 32, maximum text length of 256 tokens, LoRA rank of 8 and trained for 20,000 iterations. The temperature for the loss function is set to 0.02 and Phi-3.5-V number of sub-image crops is 4.
FSIR-CTR model was trained on our curated dataset (detailed in Section~\ref{sec:FSIR-CTR-Training-dataset}) using 4 NVIDIA A100 GPUs.
\subsection{Baseline Evaluation Protocol}
For all baseline methods (BLIP, CLIP), we employed standard dual-encoder retrieval pipelines. We first encoded all dataset images using each model's pre-trained image encoder and stored the embeddings in a vector database. For queries, we applied the text encoder with "a photo of a" prefix formatting, following established vision-language retrieval practices. Retrieval used cosine similarity to find top-k candidates from the pre-computed image embeddings. This standardized framework enables direct comparison between our few-shot methods and baseline zero-shot capabilities, isolating our techniques' contributions from implementation advantages.

\subsection{Results}
We evaluate FSIR against state-of-the-art baselines (BLIP and CLIP) on three FSIR-BD benchmark categories: FSIR-BD-Compositional-VG, FSIR-BD-Compositional-INQUIRE, and FSIR-BD-OOD.

Table~\ref{tab:FSIR-BD_Results} shows that FSIR-PL optimization achieves superior performance across all benchmarks. FSIR-PL exceeds BLIP-Large by 9.3\% on average (up to 17.9\% on OOD) and outperforms CLIP-Large by 10.5\% on average (up to 20.9\% on OOD).

FSIR-CTR-Phi-3.5-BLIP-L demonstrates an average performance gain of 3.5\% relative to BLIP-L, exhibiting superior performance on FSIR-BD-Compositional-INQUIRE and FSIR-BD-OOD, while maintaining competitive results on FSIR-BD-Compositional-VG.
FSIR-CTR-Phi-3.5-CLIP-L achieves an average improvement of 0.9\% over CLIP-L, with a notable 3.8\% enhancement on FSIR-BD-Compositional-VG, though showing slightly reduced performance on FSIR-BD-Compositional-INQUIRE.
Remarkably, FSIR-CTR-Phi-3.5 functions as a general-purpose model that was not trained on such query types and was exposed exclusively to positive reference images during training, without any negative examples.

\begin{table}[h!]
  \small
  \centering
    \setlength{\tabcolsep}{4pt} 
    \begin{tabular}{l c c c c} 
      \toprule        
      & VG & INQUIRE & OOD  & Average\\                
      \midrule        
      BLIP-L                    & \textit{41}   & \textit{28.2} &  \textit{15} & \textit{28} \\        
      BLIP-L PL                 &  \textbf{45.9}  & \textbf{33} & \textbf{32.9}  & \textbf{37.3}\\            
      FSIR-CTR-Phi-3.5-BLIP-L   & 40.8    &  \underline{29.2} & \ \underline{24.6} &  \underline{31.5} \\
      \midrule 
      CLIP-L                    & \textit{25.9}  & \underline{\textit{34.5}} & \textit{27.7} & \textit{29.4} \\
      CLIP-L PL                 &  \underline{28.3}  & \textbf{42.7} & \textbf{48.6} & \textbf{39.9} \\    
      FSIR-CTR-Phi-3.5-CLIP-L   & \textbf{29.7}   &  33 & \underline{28.2} &  \underline{30.3} \\
      \bottomrule
  \end{tabular}
  \caption{mAP retrieval performance on FSIR-BD. Values in \textit{italic} represent baseline performance for each image encoder, while values in \textbf{bold} indicate the results of best model performance and values in \underline{underline} indicate the second best results.}
  \label{tab:FSIR-BD_Results} 
\end{table}

We conducted ablation studies to evaluate the impact of the number of reference images on FSIR-PL performance, as reported in Table~\ref{tab:FSIR-BD_PL} over BLIP-L PL model. The results indicate that increasing the number of reference images consistently improves performance, with 16 reference images (the maximum tested in our experiments) yielding the best results.

\begin{table}[h!]
  \small
  \centering
  \setlength{\tabcolsep}{4pt} 
  \begin{tabular}{lcccccc}
      \toprule        
      & Baseline & 1 & 2  & 4 & 8 & 16 \\                
      \midrule        
      Compositional-VG       & \textit{41}  & 34.4 & 37.1 & \underline{41.7} & \underline{43.1} & \textbf{45.9}  \\      
      Compositional-Inquire  & \textit{28.2}  &  24.9 & 25.9 & 27.3 & \underline{28.8} & \textbf{33}  \\   
       OOD  & \textit{15} & \underline{18.7} & \underline{19.9} & \underline{21} & \underline{22.5} & \textbf{32.9}  \\  
      \bottomrule
  \end{tabular}
  \caption{mAP retrieval performance on FSIR-BD for different number of shots for BLIP-L PL. Values in \textit{italic} represent baseline performance for each image encoder, values in \underline{underline} indicate higher than baseline results, while values in \textbf{bold} indicate the results of best model performance.}
  \label{tab:FSIR-BD_PL} 
\end{table}

\section{Limitations and Future Work}
While FSIR advances the state-of-the-art in compositional and out-of-distribution image retrieval, several challenges warrant further investigation. First, the FSIR-CTR model exhibits sensitivity to reference image selection, with performance varying considerably based on the chosen examples. Second, FSIR-CTR currently utilizes only positive reference images; our preliminary experiments with negative reference images did not yield performance improvements, suggesting this remains an open research question. Third, our approach requires a pool of few-shot reference images for each retrieval task, limiting its applicability in scenarios where suitable exemplars are scarce or unavailable.

Future research should address three key directions: (1) enhancing MLLM robustness to mitigate sensitivity across diverse reference image selections; (2) developing methods to effectively leverage negative reference images for improved retrieval performance; and (3) exploring techniques for synthesizing reference images when authentic examples are unavailable. These advancements would significantly broaden the practical applicability of FSIR in real-world deployment scenarios.

\section{Conclusions}
This work makes three primary contributions to the field of image retrieval. First, we introduce FSIR-BD, a comprehensive benchmark dataset specifically designed to evaluate image retrieval using Composed queries of Text plus Reference shots (CTR). This benchmark focuses on compositional and OOD queries under challenging and realistic conditions, featuring multiple ground truth matches per query. Second, we present a novel framework with multiple approaches (FSIR-PL and FSIR-CTR) that significantly enhances retrieval performance through reference-based  inference and learning. Third, both our FSIR-PL method and FSIR-CTR are computationally efficient and enable retrieval performance optimization of existing pre-trained or custom fine-tuned image encoders, making them readily applicable to existing large-scale environments. 

\subsection*{Acknowledgements}
We acknowledge special thanks Mr. Michael Atamuk for his significant engineering contribution in developing the VidArts benchmarking system, which enabled the efficient evaluation of models, datasets, and methods. We thank Dr. Eliezer Levy, the Pnueli Lab CTO, and Lei Zhou for their continuous support throughout this project. Finally, we acknowledge the Huawei CloudBU PI Lab (particularly Zili Jin, Shuangwu Zhang, and Jackie Jing Zhang) and the Huawei CloudBU EI (particularly Ping Zhang) for their valuable support and encouragement of innovation.

\clearpage
{
    \small
    \bibliographystyle{ieeenat_fullname}
    \bibliography{main}
}

\input{appendix}

\end{document}

%% file: appendix.tex
%



\maketitlesupplementary



\section{FSIR-BD}
\subsection{FSIR-BD-Compositional-VG: Annotation process}

Notably, we initially prioritized computational approaches to reduce human labor, but with limited success. We deployed QWEN2-VL 72B, a high-performance MLLM, to determine image-query matches. After reviewing positive candidates for several queries, we observed approximately 50\% false positives, necessitating an increase in our candidate pool to 100-200 images. With the model latency of seconds per image, this approach proved prohibitively time-consuming. This attempt demonstrated insufficient performance of even top MLLM models (as of mid-late 2024) and highlighted the importance of human curation, to which we subsequently redirected efforts. The curation included three phases:

\textbf{Phase I - Initial Computation:} We encoded all 108K images as embeddings in a vector database and indexed by an approximate nearest neighbor search algorithm (ANNS). 

\textbf{Phase II - Queries definition:} We defined 50 compositional queries representing realistic scenarios. For each query, we curated the targeted HP and HN images. HP images must include all properties specified in a query, while HN images may include some aspects but never all. For example, the query \textit{"dark shirt and backpack on man in park"} requires a male adult wearing dark upper clothing, carrying a backpack, and being in a park. Corresponding HN samples include non-matching elements, such as a female, or without a backpack, or a non-park location. Notably, semantic HN may differ from computer-vision-based HNs calculated by embedding cosine similarity, a phenomenon explored later. 

\textbf{Phase III - Initial Semi-Manual Curation:} Using the vector database, we retrieved top k=1,000 candidate images per query and manually selected 50 HP and 150 HN items for each. Some queries required multiple variations to reach the 50 HP quota. This effort yielded 7,873 images, complemented with 1,200 randomly selected Flickr32K images, forming GTQR version-01 with 9,073 total images.

However, this initial version had two significant weaknesses: 1) unknown false positives (FP) due to potential errors by the single curator per query, and 2) unknown and potentially significant false negatives (FN), as negatives were checked only against their particular query.

\textbf{Phase IV – Three Rounds of FP and FN Cleanup:} We encoded GTQR version 01, and later versions 02 and 03 into a vector database, then fetched and reorganized on disk into [query]\slash true and [query]\slash false folders. "True" folders contained known HP images, while "false" folders contained CV-based HN images at a 3:1 ratio to "true" folders. This approach computationally isolated most potential false negatives and placed them alongside relevant queries. We manually reviewed folders to identify false positives and false negative in query relevant folders. After the first review round, we created GTQR-v02 and repeated the process, resulting in GTQR-v03 and again repeated the process, resulting in GTQR-v04. In our assessment, the resulting FSIR-BD-Compositional-VG GTQR-v04 is highly accurate.

\subsection{Smart Split to Test and FSR Corpuses}
To isolate the test corpus from FSR, we addressed several challenges. First, since some images matched multiple queries, random selection of positives might leave insufficient images for testing, so we prioritized images previously allocated to FSR, both in HP and HN cases. Second, to avoid homogeneity in FSR hard negatives, we divided the 16 HN into 12 closest CV-based HN and 4 farther HN. Finally, we supplemented the FSR with 100 randomly selected "easy false" images.

The FSIR-BD comprises 38,353 images and 303 queries, with 82\% allocated to the test corpus (containing an average of 37 ground truth matches per query, complemented by up to three times as many hard negatives) and 18\% of images assigned to the FSR corpus. Across sub-datasets, the average positives per query vary slightly: FSIR-BD-Compositional-VG has 55, while FSIR-BD-Compositional-INQUIRE and FSIR-BD-OOD each have 34.

\begin{figure}[ht!]
\centering
\includegraphics[width=90mm]{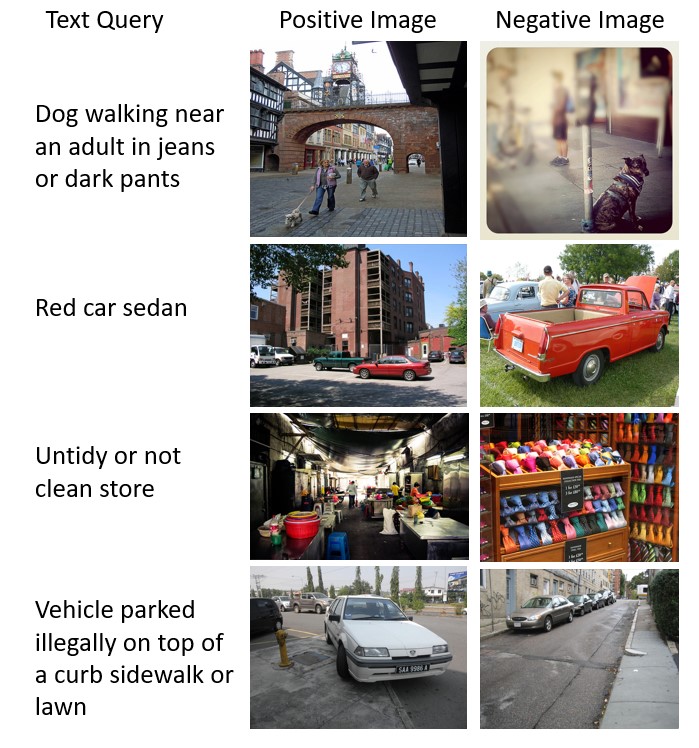}
\phantomsection
\caption{Example pf FSIR-BD-Compositional-VG data. Per text query we provide multiple positive images and negative images.}
\label{fig:vg_2}
\end{figure}

\begin{figure}[ht!]
\centering
\includegraphics[width=90mm]{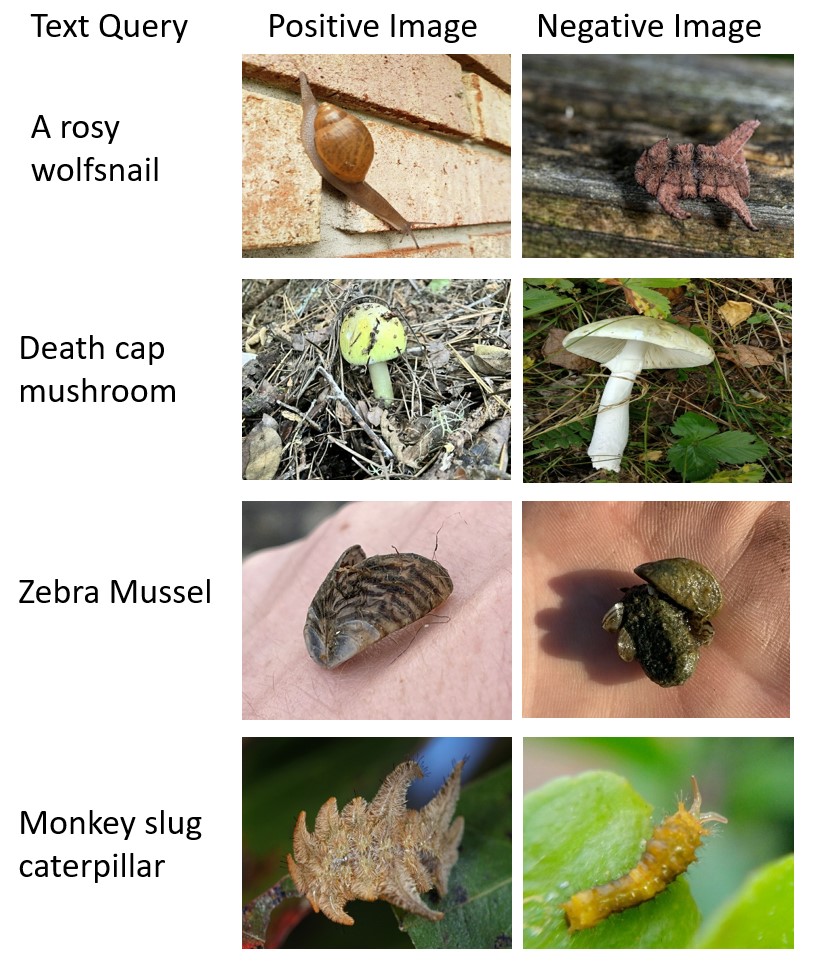}
\phantomsection
\caption{Example pf FSIR-BD-Compositional-Inquire data. Per text query we provide multiple positive images and negative images.}
\label{fig:inquire_2}
\end{figure}

\begin{figure}[ht!]
\centering
\includegraphics[width=90mm]{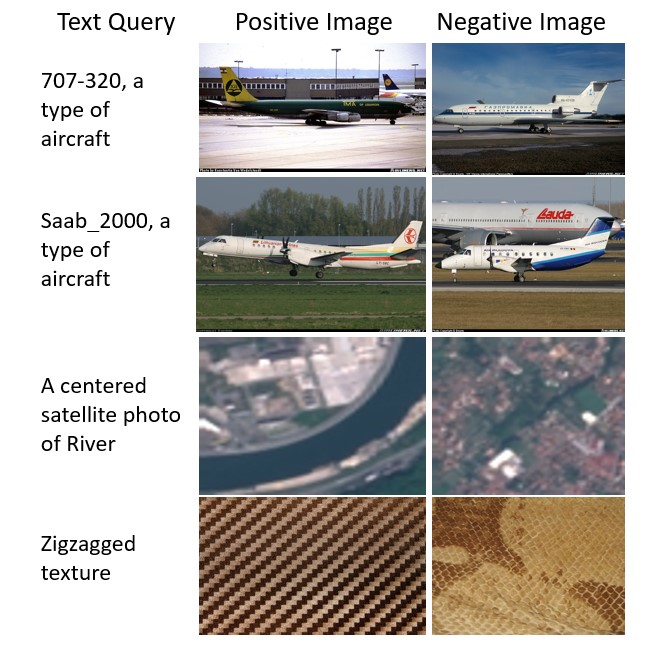}
\phantomsection
\caption{Example pf FSIR-BD-OOD data. Per text query we provide multiple positive images and negative images.}
\label{fig:ood_2}
\end{figure}

\subsection{Semantic Analysis}
We hereby provide an analysis and comparison with two leading CIR datasets, CIRR and CIRCO Table \ref{tab:FSIR-BD-Semantics}

\begin{table*}[h]
  \small
  \scalebox{0.70}{
    \begin{tabular}{llllp{13cm}}
        \toprule        
        Semantic aspect        & FSIR     & CIRR   & CIRCO   & Examples  \\                
        \midrule        
        Attributes            & 49.8\%   &        & 63.0\%  & \textit{“blond girl or woman on a wooden bench”, “red brick building with signage”, “female beautiful demoiselle”, “male Northern Elephant Seal”.} \\
        Actions               & 17.8\%   &        & 47.0\%  & \textit{“adults play with children”, “pedestrian crossing a street near park”, “macaques engaging in mutual grooming behavior”, “Dolphins performing acrobatics”.}\\
        Spatial relationships & 20.5\% & 61.4\% & 45.7\%  & \textit{“adult and child on staircase or platform or escalator”, “bus next to blue or red sign”, “bike chained illegally to a street lamp post or traffic sign but not to fence”, “grebe with babies on its back”.}\\
        Conjunctions (and, with...) & 23.4\% & 43.7\% & 75.1\% & \textit{“man and woman both wearing glasses”, “dark shirt and backpack on man in park”, “Male Blue-black Grassquit in patchy blue-black and brown plumage”}\\
        Disjunctions (or, either) & 10.6\% &      & 14.0\% & \textit{“adult and child on staircase or platform or escalator”, “window with curtains that are not red or yellow”}\\
        Negations             & 6.6\% & 11.9\%    & 12.2\% & \textit{“bike chained illegally … but not to fence”, “pedestrians … with a car that is not white”, “red brick residential building without emergency ladder”, “Caribou without palmate antlers”.}\\
        Biological taxonomy \& species-specific & 25.7\% & <1\% & <1\% & Expert queries on specific plants, animals, fungi, and even life stages or anatomical parts (\textit{“A beached Portuguese Man o’ War”, “Swallowtail butterfly caterpillar camouflaged as bird droppings”, “Japanese knotweed”}).\\
        OOD queries           & 52.5\%   & 0\%    & 0\% & 57 texture types (Examples: \textit{"knitted", "meshed", "fibrous"}), 100 Aircraft types (Examples: \textit{"MD-87", "PA-28", "757-300"}), 12 Land types as seen from satellite (Examples: \textit{"Forest", "River", "Herbaceous Vegetation"})\\
        
        \bottomrule
    \end{tabular}}
    \caption{Analysis of semantic aspects covered by FSIR-BD, CIRR and CIRCO datasets, with examples from the FSIR-BD.}
    \label{tab:FSIR-BD-Semantics}
\end{table*}

\subsection{Few-Shot Image Retrieval Training Dataset (FSIR-TD)}  \label{sec:FSIR-CTR-Training-dataset}
Training FSIR-CTR requires triplets of query text, reference image, and target image. We developed an automated pipeline to generate these triplets from image-caption pairs (based on COCO-Captions dataset).

Examples from our FSIR-TD training dataset:

\begin{figure}[ht!]
    \centering    
    \begin{subfigure}[b]{0.35\textwidth}
        \includegraphics[scale=0.7, width=\textwidth]{fsir-bd/coco_db_1.jpg}
    \end{subfigure}
    \hfill
    \begin{subfigure}[b]{0.35\textwidth}
        \includegraphics[scale=0.7,width=\textwidth]{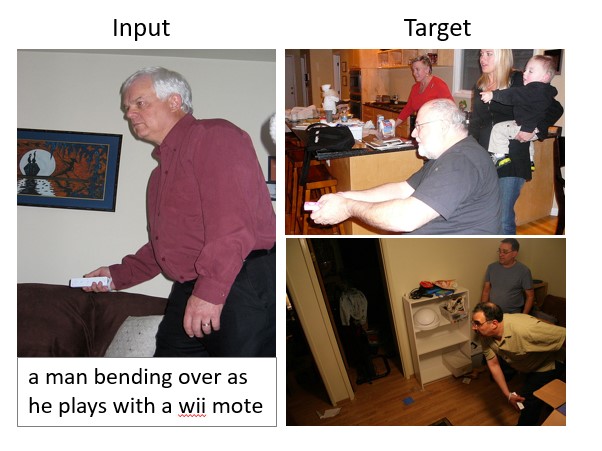}
    \end{subfigure}
    \hfill
    \begin{subfigure}[b]{0.35\textwidth}
        \includegraphics[scale=0.7,width=\textwidth]{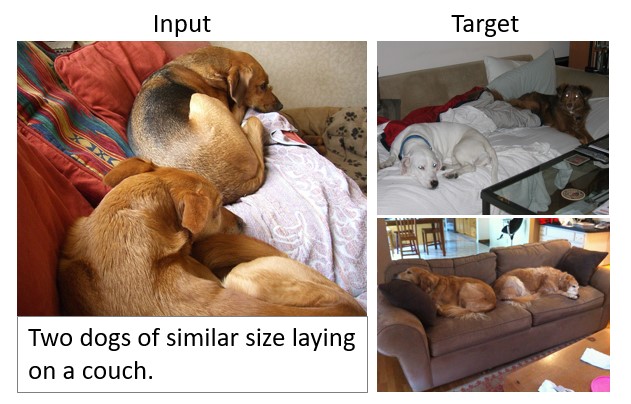}
    \end{subfigure}
    \caption{Example of FSIR-TD. Per query text we have multiple positive and hard-negative images.}
    \label{fig:fsir_td}
\end{figure}

\section{FSIR-CTR Results Analysis}

We conducted additional ablation studies on the FSIR-CTR-Phi-3.5 model. Specifically, we compared FSIR-CTR-Phi-3.5-BLIP-L against two baselines: BLIP-L and Phi-3.5-E5V. Phi-3.5-E5V is SOTA CIR model which processes images using the high-compute Phi-3.5 VLLM architecture with the same reference image selection strategy over the validation set. 

As shown in Table~\ref{tab:FSIR-BD_Results}, FSIR-CTR-Phi-3.5-BLIP-L demonstrates superior performance on the INQUIRE while achieving comparable results to BLIP-L on VG, and slightly lower than Phi-3.5-E5V on OOD benchmark . Overall, our model achieves an average improvement of 3.5\% over BLIP-L and 2\% over the Phi-3.5-E5V baseline.

\begin{table}[h!]
  \small
  \centering
    \setlength{\tabcolsep}{4pt} 
    \begin{tabular}{l c c c c} 
      \toprule        
      & VG & INQUIRE & OOD  & Average\\                
      \midrule        
      BLIP-L         & \textit{41}   & \textit{28.2} &  \textit{15} & \textit{28} \\              
      Phi-3.5-E5V       & 31.4  & 28.6  &  \textbf{28.6}  & \underline{29.5} \\    
      FSIR-CTR-Phi-3.5-BLIP-L   & \underline{40.8}    &   \textbf{29.2} & \ \underline{24.6} &  \textbf{31.5} \\    
      \bottomrule
  \end{tabular}
  \caption{mAP retrieval performance on FSIR-BD. Values in \textit{italic} represent baseline performance for each image encoder, while values in \textbf{bold} indicate the results of best model performance and values in \underline{underline} indicate the second best results.}
  \label{tab:FSIR-BD_Results} 
\end{table}